\documentclass[letterpaper, 10pt, conference]{ieeeconf}
\IEEEoverridecommandlockouts 
\usepackage{graphicx} 
\usepackage{textcomp}
\usepackage{tikz}
\usetikzlibrary{matrix,arrows,chains,calc,positioning}
\usepackage{lipsum,adjustbox} 
\usepackage[normalem]{ulem}
\usepackage{cite}
\usepackage{flushend} 
\def\algbackskip{\hskip-\ALG@thistlm}
\usepackage{amsfonts}
\usepackage{caption}
\usepackage{subcaption}
\usepackage{soul}
\def\rebuttal#1{{\color{black}{{#1}}}}
\usepackage{todonotes}
\usepackage[utf8]{inputenc}
\usepackage{pstricks}
\usepackage{url}
\begin{document}
\title{\LARGE \bf Close-Proximity Underwater Terrain Mapping Using Learning-based Coarse Range Estimation}
\author{Bilal Arain$^{\dagger\ddagger}$, Feras Dayoub$^{\dagger\star}$, Paul Rigby$^\ddagger$, Matthew Dunbabin$^{\dagger\star}$ \thanks{$^\dagger$ Queensland University of Technology (QUT), 2 George Street, Brisbane, QLD 4000, Australia. 
{{\tt\small arain.bilal@gmail.com}.}}
\thanks{$^\ddagger$ The Australian Institute of Marine Science, PMB3, Townsville MC, QLD, Australia. {{\tt\small p.rigby@aims.gov.au}.}}
\thanks{$^\star$ The Australian Research Council Centre of Excellence for Robotic Vision (ACRV). {{\tt\small feras.dayoub@qut.edu.au},} {{\tt\small m.dunbabin@qut.edu.au}.}}}

\thispagestyle{empty} \maketitle

\begin{abstract} 
This paper presents a novel approach to underwater terrain mapping for Autonomous Underwater Vehicles (AUVs) operating in close proximity to complex 3D environments.  The proposed methodology  creates a probabilistic elevation map of the terrain using a monocular image learning-based scene range estimator as a sensor. This scene range estimator can filter transient objects such as fish and lighting variations. The mapping approach considers uncertainty in both the estimated scene range and robot pose as the AUV moves through the environment. The resulting elevation map can be used for reactive path planning and obstacle avoidance to allow robotic systems to approach the underwater terrain as closely as possible. The performance of our approach is evaluated in a simulated underwater environment by comparing the reconstructed terrain to ground truth reference maps, as well as demonstrated using AUV field data collected within in a coral reef environment. \rebuttal{The simulations and field results show that the proposed approach is feasible for obstacle detection and range estimation using a monocular camera in reef environments.}
\end{abstract}

\section{Introduction}
\label{sec:introduction}
Dependable navigation of Autonomous Underwater Vehicles (AUVs) \rebuttal{in close proximity (i.e. within $0.25$--$1\mathrm{m}$) to complex 3D terrain is important for applications such as scientific image based surveys} of the environment (such as coral reefs) and engineering inspections. Successful completion of these missions demands that the AUV can approach obstacles closely, while ensuring that it does not actually collide with them. A pre-requisite to this behaviour is the robust perception and local mapping of obstacles and surroundings, which is the focus of this paper. Traditional acoustic sensors for underwater mapping (e.g. sonars) become less reliable in very shallow water marine environments due to distortion, scattering and minimum blanking distances \cite{mmb15}. Thus, a reliable visual perception system is a viable option to detecting and avoiding obstacles at close range and with high precision. 

\begin{figure}
    \centering
    \includegraphics[width=0.48\textwidth]{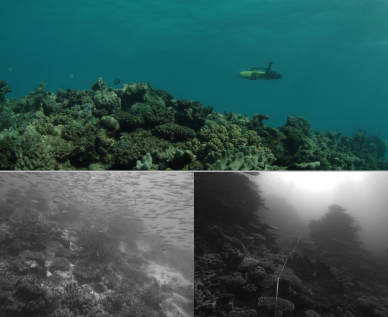}
    \caption{Examples of challenges for AUV navigation in close-proximity to coral reef environments. (Top) A small AUV traversing at 1.0 m above a relatively flat coral reef. (Lower left) An image from the AUV showing cluttered and dynamic (fish) obstacles. (Lower right) An image showing strong lighting variations.}
    \label{fig:operating_challenges}
\end{figure}

Robust feature detection using stereo-vision can be difficult due to constraints such as light absorption~\cite{rdg14}, transients such as marine life, background sun glare and motion blur~\cite{dcb04}. Figure~\ref{fig:operating_challenges} shows some examples of challenging coral reef environment for AUV navigation in close proximity to the complex 3D terrain. We have shown in~\cite{amr19} that these factors can lead to falsely identified objects or free space, which may prevent robust terrain-based navigation. These limitations prevent stereo-based approaches from being used for real-time obstacle detection~\cite{rpm14} and are primarily used for offline post-processing of the images for habitat classification~\cite{wpw10,wpj12}.

In this paper we present a new approach to underwater terrain mapping using a coarse learning-based scene range estimator. We have approximated the problem of continuous range regression with a discrete range estimation problem using a monocular camera.  Building on our previous work~\cite{amr19}, we segment and classify the monocular images observed by the AUV into discrete obstacle ranges and show that this approach is resilient to dynamic objects (e.g. fish) and visibility aberrations. The predicted range of each image pixel, with associated uncertainty, is then used as a sensor model to construct the underwater terrain map. This probabilistic elevation map of terrain is evaluated in simulations and with field data.

The remainder of the paper is structured as follows; Section~\ref{sec:related_work} summarises related research in monocular-image based range estimation for classification and underwater mapping approaches for AUV applications. Section~\ref{sec:methodology} describes the proposed methodology for terrain map generation using the image-based classification and uncertainty in range estimation. Section~\ref{sec:results} presents results and performance analysis using field data collected during coral surveys and \rebuttal{underwater} simulations, with conclusions in Section~\ref{sec:conclusions}.

\section{Related Work}
\label{sec:related_work}
Underwater terrain mapping in close proximity to obstacles requires an AUV to perform robust obstacle detection and avoidance in real-time. Visual perception and terrain mapping of coral reefs can be achieved by constructing a dynamic representation of the scene in terms of occupancy grids or digital elevation maps. In this paper, the relevant literature is divided into: (a) Supervised monocular image-based range estimation through classification; (b) Vision-based underwater terrain mapping for navigation.

\subsection{Supervised Monocular Image-based Range Estimation through Classification}
A number of techniques have been used for estimating the scene range from monocular cameras in the computer vision literature~\cite{bhoi19}. An early method to produce dense pixel depth estimates~\cite{epf14} used a data driven approach to learn features using a two scale network. This work has been extended by using a \rebuttal{Conditional Random Field (CRF)} model for the scene depth regularization~\cite{lsd15}, using multi-class classification loss~\cite{cws17} and residual learning with reverse Huber loss function~\cite{lrb16}. In~\cite{fgw18},  continuous scene range is transformed into a discrete number of intervals and a network is trained by ordinal regression loss to predict the range distribution. Recently, a discrete classification problem is formulated in~\cite{yhr19} to predict a valid range Probability Density Function (PDF) by learning the model as an independent binary classifier. This approach regularizes the model and gives better uncertainty estimates and scene depth predictions.

Although pixel-wise classification can be used to estimate scene depth, there have been limited examples of range estimation for underwater obstacle avoidance using a monocular camera.  Inspired by the work in~\cite{msu18}, we formulate the underwater discrete range prediction as a multi-modal semantic segmentation problem. Whilst~\cite{msu18} predicts spatially dense range, in our method each bin covers a fixed value of a range and the labels of the bins are defined according to the range.

\subsection{Vision-based Underwater Terrain Mapping for Navigation}
Underwater terrain-based navigation approaches primarily focus on the localization task by estimating the state of the vehicle using on-board sensors~\cite{mm17}. \rebuttal{Approaches such as Simultaneous Localization And Mapping (SLAM) build a map of the environment using a set of features identified within sonar or vision data measurements~\cite{pss14}. However, loop-closure is required to minimize the error in the robot position~\cite{wpw10} which can be challenging for AUVs to achieve when conducting low-altitude coral survey tasks. Moreover, the estimation process becomes computationally expensive to solve in real-time~\cite{mm17} as the map size increases. A robot-centric approach, where a map frame is defined with respect to the robot, has the advantage of resilience to drift in the robot pose. The pose uncertainty estimate can be propagated to the map data resulting in a consistent representation of the terrain sufficient for local path planning and obstacle avoidance. Additionally, with smaller fixed size local maps, the computations are deterministic and could be achieved in real-time.}

Monocular vision-based reactive avoidance methods are proposed in~\cite{rpm14, mtr15, dhe16}  using visual perspective invariants. However, these approaches prioritise `escaping' from the obstacles including marine life, while in this work we create a probabilistic elevation map of the terrain, which will enable the risk of collision to be managed. Conceptually, our approach is similar to~\cite{fankhauser18}, where the terrain mapping is used for local representation of the robot's surroundings while incorporating range estimation uncertainty to update the map. To quantify the errors in the model, epistemic and aleatoric uncertainty metrics have been used in the literature~\cite{kg17}. We have considered a multi-modal distribution over estimated ranges from the semantic classifier to use as an uncertainty estimation as presented below. 

\begin{figure}
    \centering
    \includegraphics[width=0.48\textwidth]{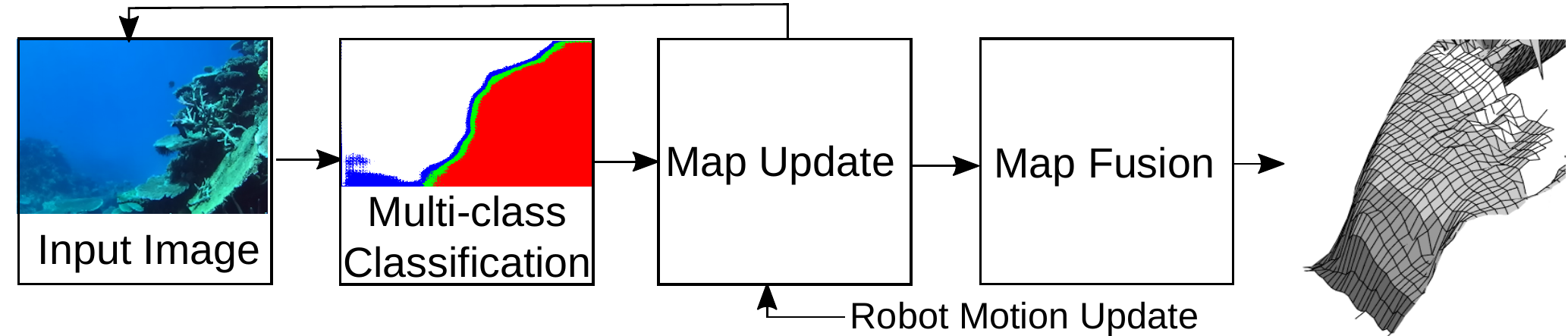}
    \caption{Schematic overview of the terrain map generation. The multi-class classifier generates coarse range estimate of the terrain from the monocular camera in field-of-view. A probabilistic terrain map is created using the range measurements and the robot motion update. The estimated height along with the confidence bounds are computed to produce the fused map.}
    \label{fig:system_overview}
\end{figure}

\section{Terrain Mapping using Learning-based Coarse Range Estimation}
\label{sec:methodology}
A schematic overview of our approach is shown in Figure~\ref{fig:system_overview}. The proposed method for terrain elevation mapping infers the range class labels in a monocular image sequence using a \rebuttal{Deep Convolutional Neural Network (DCNN)}.  The classifier is trained to predict the obstacles in the monocular scene as a discrete set of ranges/distance (near, mid-field, far, or no obstacle) across the image. The predicted range of each image pixel along with its uncertainty then forms a ‘sensor model’, along with the robot motion update. \rebuttal{It is assumed that an estimate of the robot pose is available using other on-board sensors (e.g. IMU, pressure) or by using visual odometry methods~(see Section~\ref{sec:experimental_setup})}. A corresponding point cloud of the predicted ranges along with their associated uncertainty are then transformed to the corresponding height measurements with confidence bounds. The following sections describe the approach for probabilistic underwater terrain (elevation) mapping using the semantic classes developed and used within this paper.

\subsection{Terrain Map Generation}
\label{sec:map_generation}
Due to the inherent difficulties of obtaining globally accurate position information without sophisticated externally introduced localisation systems, we are inspired by the work of Fankhauser~\cite{fankhauser18} to generate \emph{local} maps of the environment as the AUV traverses through it expressed as a \emph{robot-centric} formulation.

\begin{figure}
    \centering
    \def\svgwidth{0.48\textwidth}
    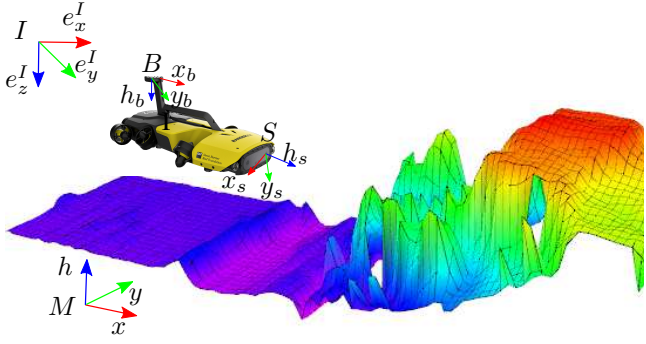
    \caption{Illustration of the coordinate systems used for the terrain (elevation) mapping. The inertial reference frame  $I$ is in commonly used NED coordinates, with the local right-hand map coordinate frame $M$ ($x, y, h$) selected to an arbitrarily selected location, and the fixed (sensing) frame $S$ ($x_s, y_s, h_s$) attached to the AUV.}
    \label{fig:coordinate_system_overall}
\end{figure}

The coordinate system used for terrain mapping is illustrated in Figure~\ref{fig:coordinate_system_overall}. The four coordinate frames are defined with an inertial frame~\emph{I} in \rebuttal{North-East-Down (NED)} coordinates  typical for underwater environments, the base frame~\emph{B} that moves with the robot, a map frame~\emph{M} relative to the base frame, and a sensor frame~\emph{S} attached to the camera on the robot. The unit vectors $\{e_{x}^{I},e_{y}^{I},e_{z}^{I}\}$ are along the respective inertial axes. It is assumed that there are known translation and rotation transformations ($\mathbf{r}_{BS},\phi_{BS}$) between the base~\emph{B} and the sensor \emph{S} frames. The base frame~\emph{B} is related to the inertial frame \emph{I} through the translation~$\mathbf{r}_{IB}$ and rotation $\phi_{IB}$. \rebuttal{The Euler angle parametrization and the convention ``yaw-pitch-roll'' is used for the rotation matrix $\phi_{IB}$.} 

The local map is updated at each time step using the range estimation from a single image collected in the sensor frame~$S$ and transformed to the map frame~$M$. At each measurement step, a cell at ($x,y$) in the local map frame is denoted by $P_{xy} = (x, y, \hat{h})$ where $\hat{h}$ is the estimated height. To obtain the estimated height of the cell from the measured height $\tilde{h}$, the measurement is approximated by a Gaussian probability distribution $\tilde{h}\sim \mathcal{N}\left(h_{c}, \sigma^{2}_{h_{c}}\right)$ with mean $h_{c}$ and variance $\sigma^{2}_{h_{c}}$. The variance of the height measurement within the map $\sigma^{2}_{h_{c}}$ is given by:

\begin{eqnarray}
\label{app_eq:2}
\sigma^2_{h_{c}} = J_{S} \ \mathrm{cov} (S) \ J^T_S + J_\phi \ \mathrm{cov} (\phi_{IS}) \ J_{\phi}^{T}
\end{eqnarray}
where $\mathrm{cov}(S)$ is the range prediction covariance matrix of the range sensor model~(see Section~\ref{sec:uncertainty}) and $\mathrm{cov}(\phi_{IS})$ is the covariance matrix of the sensor rotation being a sub matrix of the $6\times 6$ pose covariance matrix. The Jacobians for the range predictions \rebuttal{($J_{S}$)} and the sensor frame rotation \rebuttal{($J_{\phi}$)} are determined by

\begin{eqnarray}
\label{eq:jacobians}
J_{S} = \frac{\partial h_{c}}{\partial_S \mathbf{r}_{SP}}, \mathrm{and}\quad \quad
J_{\phi} = \frac{\partial h_{c}}{\partial \phi_{SM}}.
\end{eqnarray}
The height measurement is transformed into the map frame using the position of the measurement $_S\mathbf{r}_{SP}$ in the sensor frame such that
\begin{eqnarray}
\label{eq:mean_height}
h_{c} = \mathbf{P}\left(\phi_{SM}^{-1}\left(_S\mathbf{r}_{SP}\right)-_M\mathbf{r}_{SM}\right),
\end{eqnarray}
where $\mathbf{P}=[0\, 0\, 1]$ is the projection matrix to map the scalar height measurement in the map frame. The measurement and variance ($h_{c}, \sigma^{2}_{h_{c}}$) are then fused with the existing map estimate ($\hat{h},\sigma^{2}_{h}$) by a one-dimensional Kalman filter given by

\begin{eqnarray}
\label{eq:kalman_filter}
\hat{h}^{+} = \frac{\sigma^{2}_{h_{c}} \hat{h}^{-} + \hat{\sigma}^{2-}_{h} \tilde{h}}{\sigma^{2}_{h_{c}} + \hat{\sigma}^{2-}_{h}}, \quad \quad 
\sigma_{{h}}^{2+} = \frac{\hat{\sigma}_{{h}}^{2-} \sigma^{2}_{h_{c}}}{\hat{\sigma}_{{h}}^{2-} + \sigma^{2}_{h_{c}}},
\end{eqnarray}
where estimates before and after an update are denoted by a $-$ superscript and a $+$ superscript respectively. A final map fusion step can be performed within the elevation mapping framework by computing the mean height $(\hat{h}_{i}, h_{i_{min}}, h_{i_{max}})$ along with the confidence bounds for every cell $i$ as the weighted mean from all cells within $2\sigma$ confidence ellipse in the $x-y$ plane of a cell. 

To obtain the fused terrain map using~(\ref{eq:kalman_filter}), the values of the range prediction covariance matrix, $\mathrm{cov(S)}$ in~(\ref{app_eq:2}), are required of the range sensor model. In this work, the DCNN presented discussed in the following sections  is considered as the ‘sensor’  for this covariance.

\begin{figure}
    \centering
    \includegraphics[width=0.48\textwidth]{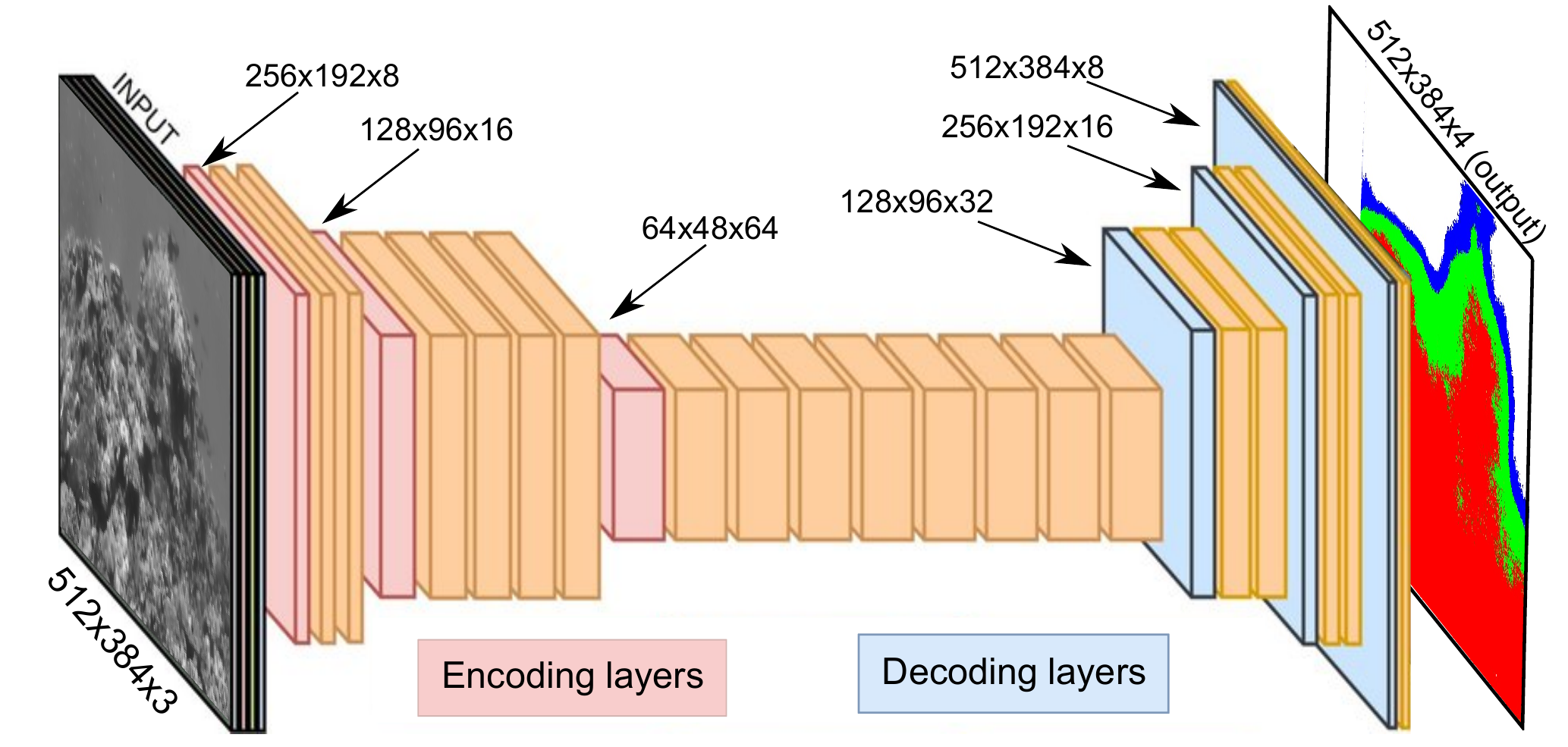}
    \caption{The ERFNet DCNN architecture used in this work for course scene range estimation. An example input image and its corresponding output image are the same resolution ($512\times384$).}
    \label{fig:network_architecture}
\end{figure}

\subsection{Coarse Range Estimation using Monocular Image Semantic Segmentation}
\label{sec:seg_models}

The goal of our approach is not necessarily to faithfully reproduce fine-scale topology of the terrain, but to provide a framework which gives confidence bounds on the terrain relative to the robot for safe navigation when in close proximity to obstacles.

The detection of obstacles and their range from the camera is based on our previous work using semantic image segmentation~\cite{amr19}. In this work, a semantic segmentation approach is used to learn obstacles from a 2D training image set. Obstacle segmentation is considered as a multi-class, $C + 1$, problem where obstacles are sub-divided into $C$ classes based on their distance from the camera with everything else considered not be an obstacle.

Using the Bonnet toolkit~\cite{ms18}, a model based on the variable receptive field non-bottleneck concept of ERFNet~\cite{rab18} is employed. This is achieved through factorized convolutions of diverse receptive fields. In this work, $4$ convolutional layers of increasing complexity, K = [$8$; $16$; $32$; $64$], are used as illustrated in Figure~\ref{fig:network_architecture} to produce a segmented obstacle range image. \rebuttal{Full images are passed to the network with the output being a pixel wise prediction of obstacle range consisting of $C = 3$ obstacle classes based on their distance from the camera. These classes correspond to near ($\tau_m$--$2\mathrm{m}$), mid-field ($2$--$3\mathrm{m}$), far ($3$--$4\mathrm{m}$) and a fourth class being free-space ($>\!4\mathrm{m}$). The minimum effective detection distance is $\tau_m\!=\!45\mathrm{cm}$. The training data was generated by manual segmentation and classification of images, and it was found through trial and error that humans could infer range reliably with this number of obstacle classes.}

In this work, we use the semantic image segmentation to obtain course range and height estimates of the terrain in front of the robot. At each time instance, a semantically labelled image is produced that predicts the scene regions within each class range referenced to the sensor frame. Using the intrinsic camera parameters, each pixel $u,v$ in each class is transformed to an elevation (${h}_{c_j}$) in the map frame using~(\ref{eq:mean_height}) for each local map cell $i$. The uncertainty of the class prediction is also used to update the associated map cell. 

\begin{figure}
    \centering
    \includegraphics[width=0.48\textwidth]{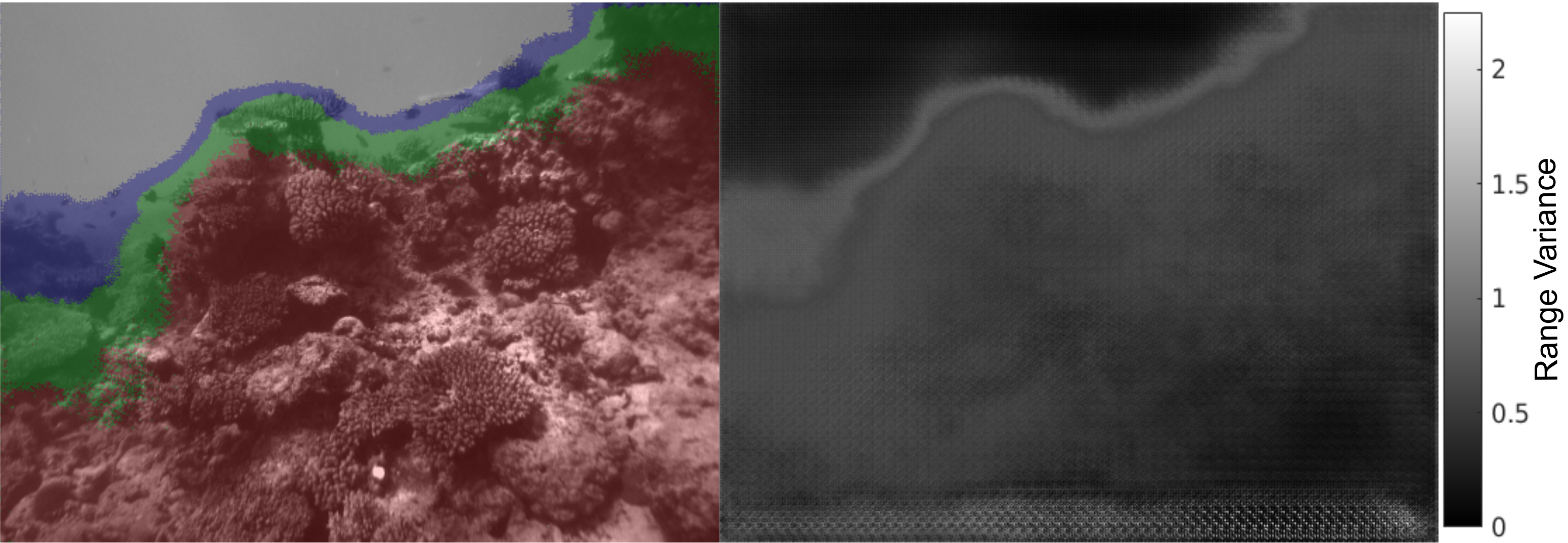}
    \caption{(Left) Original reef image with predicted binned ranges overlaid for comparison, and (Right) the calculated variance, normalised and converted to a grayscale image for visualisation. The colors represent the predicted obstacle range based on training images (i.e. red is near, green is mid-field, blue is far and free-space is white). The range variance is scaled from zero (black) to maximum $2.25$ (white). Note that the free space class range is set to $5\mathrm{m}$, which gives us the maximum mean range of $3.5\mathrm{m}$ and the maximum range variance of $2.25$.}
    \label{fig:variance_calculation}
\end{figure}

As the scene is classified into four range classes and the exact topology between the class boundaries is not known nor available (assuming only a monocular camera), it is difficult to determine the exact height to update the local map within each class boundary. It can be hypothesised that in a typical underwater scene, the class boundary is at its closest to the actual range bin value. Therefore, we update the height in the local map for only those pixels along the top edge of the predicted class boundaries.  An evaluation of these range estimation and map update cases is given in Section~\ref{sec:results}. The sensor covariance matrix is obtained from the class uncertainty in the predicted ranges as discussed below.

\subsection{Class Uncertainty Prediction}
\label{sec:uncertainty}
The probabilistic mapping framework employed in this paper (see Section~\ref{sec:map_generation}), requires an estimate of the uncertainty in the coarse range predictions from the DCNN. To obtain the class prediction uncertainty, the distribution of predicted classes are used from the output of the network as illustrated in Figure~\ref{fig:network_architecture}. We use variance in the prediction of our deep neural network as a range sensor model uncertainty to improve the estimated height of each cell in generating the terrain map. To quantify the uncertainty using the class probability distribution from the classifier, the range mean and variance are computed by

\begin{eqnarray}
\label{eq:range_mean_var}
\mu_{z} = \sum_{i=1}^{4} r_{z_{i}}P(r_{i}),\quad 
\sigma^{2}_{z} = \sum_{i=1}^{4}(r_{z_{i}}-\mu_{z})^{2}P(r_{i}),
\end{eqnarray}
where $r_{z_{i}}$ is the predicted range of class $i$, $P(r_{i})$ is the corresponding probability distribution of the predicted classes, $\mu_{z}$ and $\sigma^{2}_{z}$ are the mean and variance of the probability distribution for each pixel in the resulting segmented output image respectively. 

In contrast to previously used range sensor models~\cite{fankhauser18,nil12}, we propose the covariance of the range sensor model to be $\mathrm{cov(S)}=[0\, 0\, \sigma^{2}_{z}]$ in~(\ref{app_eq:2}) which is a $3\times3$ diagonal matrix and is updated for each pixel of the resulting segmented image over an estimated range class. Figure~\ref{fig:variance_calculation} illustrates an example class uncertainty estimation for an input image.  

\begin{figure}
    \centering
    \includegraphics[width=0.48\textwidth]{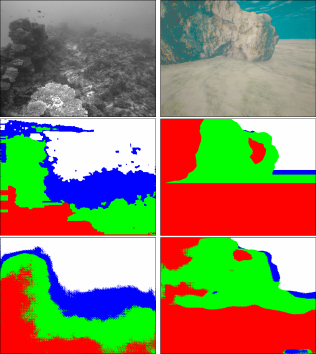}
    \caption{Example results of image segmentation for a scene from a real coral reef (left column), and the simulated \rebuttal{underwater environment} (right column). The top row is the input image to the classifier, the middle row is the ``true" binned range estimate obtained from a stereo camera, and the lower row is the prediction from a monocular camera. The colors represent the predicted obstacle range based on training images (i.e. red is near, green is mid-field, blue is far and free-space is white).}
    \label{fig:segmentation_real_sim}
\end{figure}

\section{Experimental Setup}
\label{sec:experimental_setup}
\begin{figure}
 \centering
 \includegraphics[width=0.48\textwidth]{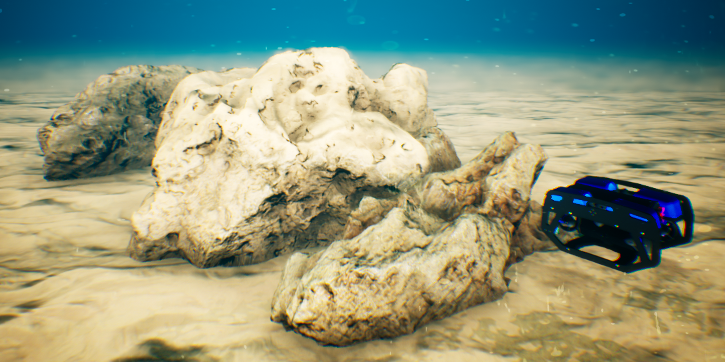}
 \caption{An example 3D scene from the \rebuttal{underwater} simulator used for evaluating the probabilistic terrain mapping methodology. The robotic system is shown as a scale guide ($0.5\mathrm{m}\!\times\!0.5\mathrm{m}$) and is pointed in the direction of travel over the 3D structure.}
 \label{fig:simulation_environment}
\end{figure}

To facilitate evaluation of the algorithms and approaches within this paper, a custom underwater simulation environment was developed within Microsoft AirSim~\cite{sdl18}. The simulator is built using various off-the-shelf Unreal Marketplace components to create an underwater lighting model and scenes of rocky 3D underwater structures suitable for complex terrain mapping. A model of a small \rebuttal{Remotely Operated underwater Vehicle (ROV)} was integrated with a custom camera model similar to that used on AUVs. Example images of a simulated 3D scene and images from the AUV's simulated camera are shown in Figures \ref{fig:segmentation_real_sim} and \ref{fig:simulation_environment}.

To evaluate the performance of the resulting terrain map using the proposed methodology, stereo image datasets were collected during coral reef surveys at John Brewer reef in the Great Barrier Reef. The images were collected using the RangerBot AUV~\cite{ddl19}, shown in Figure~\ref{fig:coordinate_system_overall}. \rebuttal{It is assumed in the experiments that the AUV pitch is stabilized, however this is not a strict condition as the map frame is with respect to the base frame of the AUV which includes attitude of the platform. The AUV pose was estimated by fusing the visual odometry estimation from the downward stereo camera and the on-board IMU. All image processing and data capturing runs on-board the  AUV using an NVIDIA Jetson TX2 module. Full details of the pose estimation methods used for the RangerBot AUV are contained in~\cite{ddl19}.}  In order to train the proposed classifier for what is considered ‘appropriate’ obstacles for the coral survey tasks, the AUV captured images from the front stereo pair with the AUV manually guided over suitable locations and around the complex reef terrain. 

\begin{table}[bp]
\renewcommand{\arraystretch}{1.5}
\caption{Coarse Range Estimation Performance.}
\centering
\footnotesize
\begin{tabular}{|c|c|c|c|c|c|c|c|c|}
\cline{1-5}
\multicolumn{5}{|c|}{Simulations}\\
\cline{1-5}
\multicolumn{1}{|c|}{Number of samples} & \multicolumn{2}{|c|}{Mean Accuracy} & \multicolumn{2}{|c|}{Mean IoU}\\
\cline{1-5}
Training & Validation & Test & Validation & Test \\
\hline
284 & 0.935 & 0.777 & 0.753 & 0.474\\
\cline{1-5}
\multicolumn{5}{|c|}{Experiments}\\
\cline{1-5}
\multicolumn{1}{|c|}{Number of samples} & \multicolumn{2}{|c|}{Mean Accuracy} & \multicolumn{2}{|c|}{Mean IoU}\\
\cline{1-5}
Training & Validation & Test & Validation & Test \\
\hline
50 & 0.844 & 0.791 & 0.671 & 0.627\\
\cline{1-5}
\end{tabular}
\label{tab:data_performance}
\end{table}

\vspace{-5pt}
\section{Results}
\label{sec:results}
\subsection{Coarse Range Estimation}
\label{sec:range_estimation_results}
Two networks were trained for this work; one using the simulator generated images, and another using field data collected on the Great Barrier Reef. \rebuttal{For the simulation analysis, a custom underwater environment was created using the Unreal Game Engine. Within this environment, an underwater robot with cameras was simulated to generate an image sequence as the robot moved through the scene (see Figure~\ref{fig:simulation_environment}). 
\begin{figure}[tp]
 \centering
 \begin{subfigure}[b]{0.5\textwidth}
 \centering
 \includegraphics[width=\textwidth]{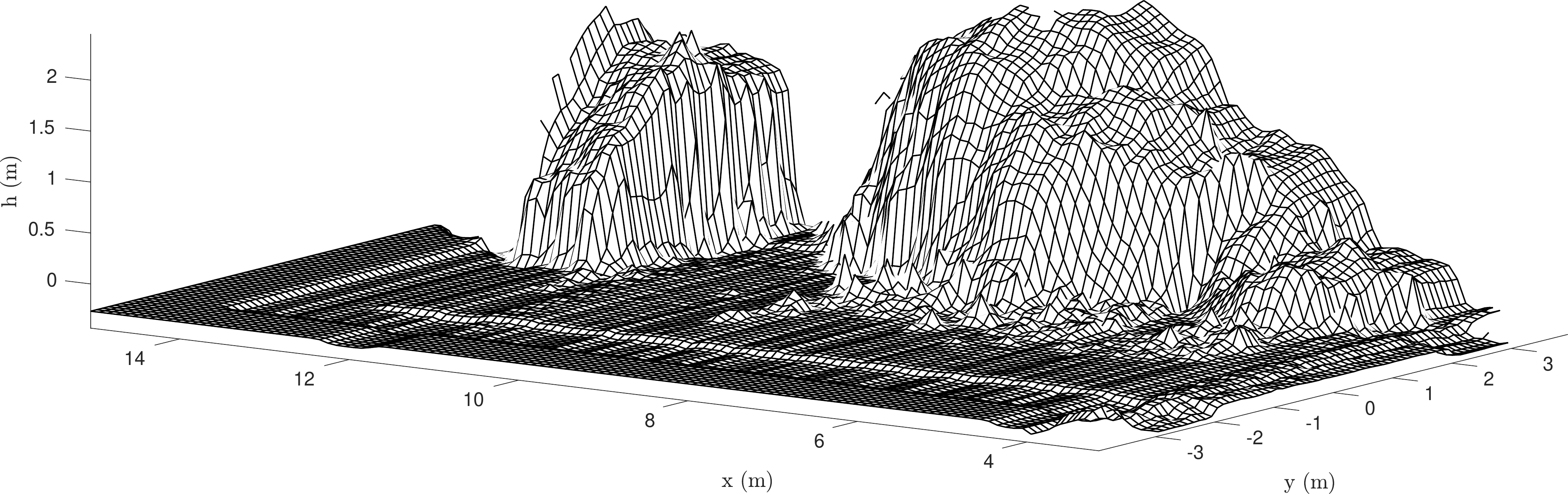}
 \caption{Example of 3D occupancy grid of the simulator terrain used as a ground truth for algorithm evaluation.}
 \label{fig:contour_true_map}
 \end{subfigure}\\
 \begin{subfigure}[b]{0.5\textwidth}
 \centering
 \includegraphics[width=\textwidth]{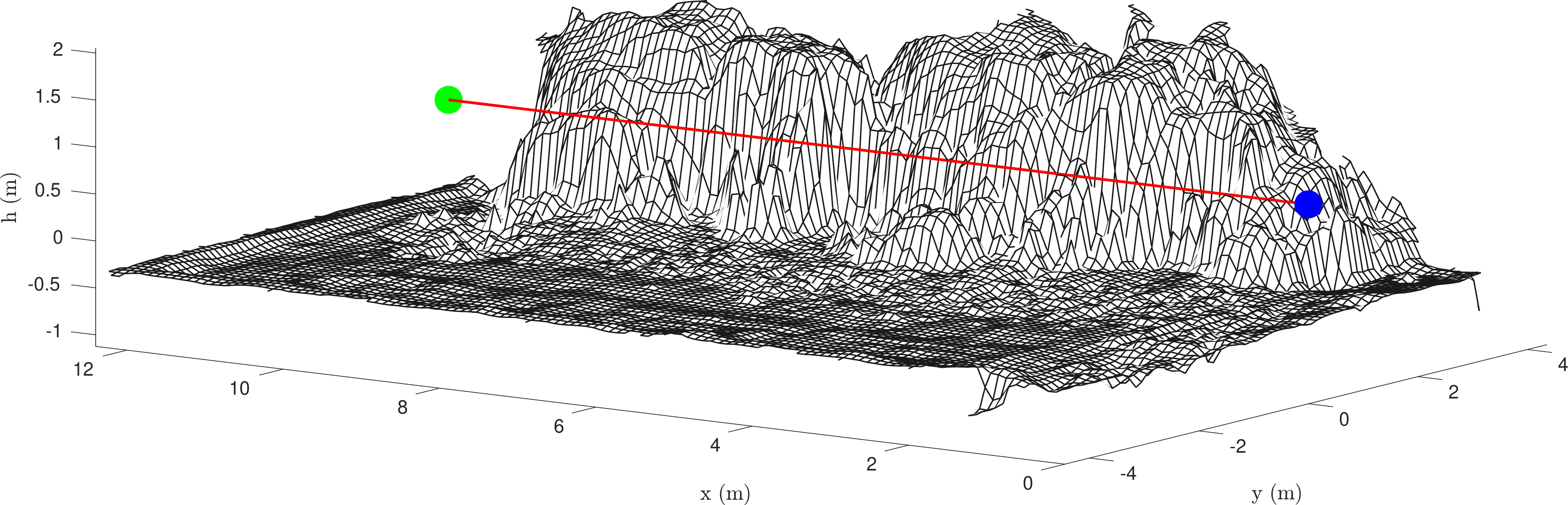}
 \caption{Example of the predicted map when accounting for the classification uncertainty. The robot traversed across the transect (red) at a constant~$h\!=\!1\mathrm{m}$ from start point (blue marker) to end point (green marker) along the x-axis, while y-axis was set to zero.}
 \label{fig:contour_fused_map}
 \end{subfigure}\\
 \begin{subfigure}[b]{0.5\textwidth}
 \centering
 \includegraphics[width=\textwidth]{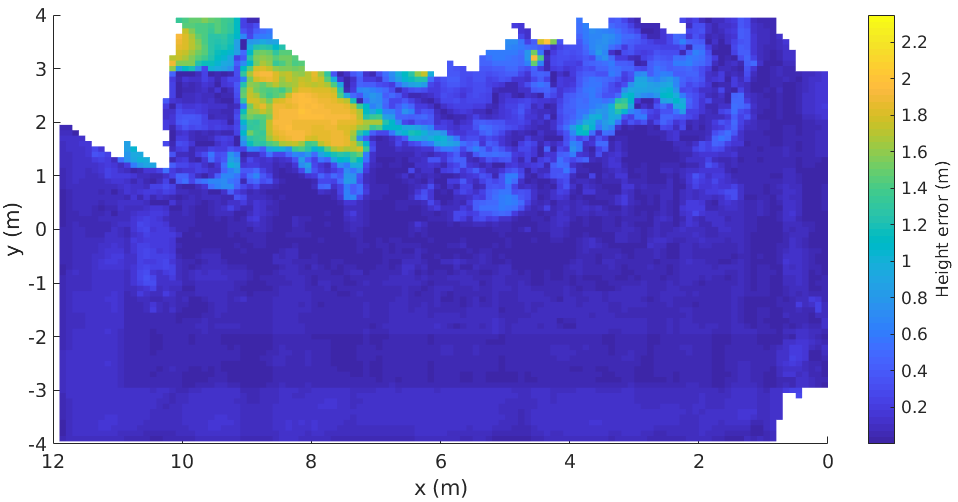}
 \caption{Error between the predicted terrain map compared to the ground truth map (top) to quantify the classifier performance.}
 \label{fig:contour_error_map}
 \end{subfigure}
 \caption{Example of the simulation~$1$ scenario for an AUV to follow the transect line in close proximity to the unknown terrain.}
 \label{fig:contour_map}
\end{figure}
This simulation framework allows artifacts such as light attenuation, surface glint and glare; however at this stage spurious features such as fish and other marine life is not modelled.} Using this simulator, annotated training images were generated by ray-tracing each pixel in the simulated camera image to the obstacle to obtain its range then discretizing the range into respective classes. \rebuttal{Training of the field model followed a similar approach with the range classes obtained by stereo matching such as LIBELAS~\cite{gru10}. Interested readers can refer to~\cite{amr19} for comparison between the coarse range estimation and stereo matching approaches.} A total of $807$ simulation annotated images and $50$ field annotated images were used to train the respective models. Figure~\ref{fig:segmentation_real_sim} shows examples of a training image, their corresponding class annotations, and the predicted segmentation for both a real (field) and a simulated scene. Table 1 summarizes the quantitative performance of the coarse range estimation in terms of the mean accuracy of all classes and the Intersection over Union (IoU) metrics. The models were trained with images of size $512\!\times\!384$ pixels using the median frequency loss with a learning rate of $10^{-3}$. The tuning parameters $\gamma$~\cite{lgg18} and $\epsilon$~\cite{kb14} were set to $2$ and $10^{-8}$ respectively.

\begin{figure}
    \centering
    \includegraphics[width=0.48\textwidth]{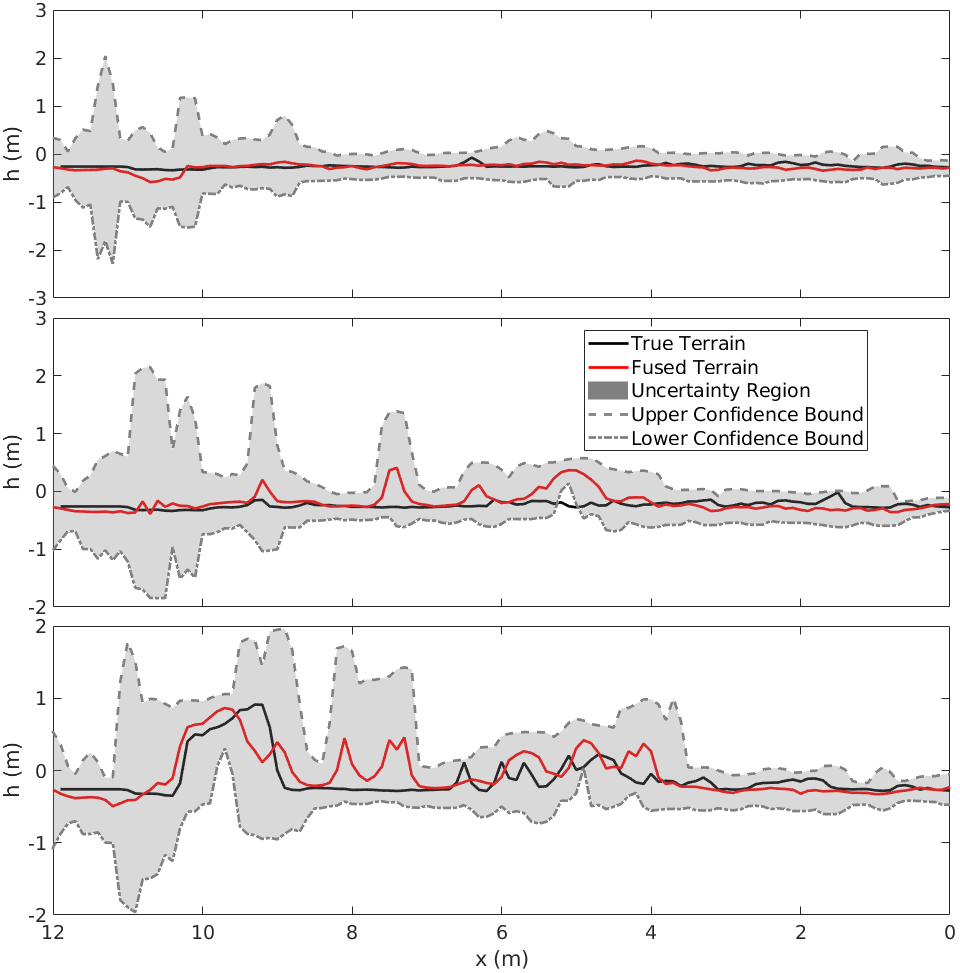}
    \caption{The cross section view of the contour terrain map in the vehicle direction at $y\!=\!0\mathrm{m}$ (Top), at $y\!=\!0.5\mathrm{m}$ (Middle), and at~$y\!=\!1\mathrm{m}$ (Lower) in the y-axis showing the comparison between resulting predicted terrain and the true terrain along with the estimated upper and lower confidence bounds. Note that the ground truth offset height has been shifted to the initial robot-centric pose. }
    \label{fig:Sim_cross_section_comparison}
\end{figure}

\subsection{\rebuttal{Evaluation in Simulated 3D terrain}}
\label{sec:simulation}
To evaluate the proposed methodology, two simulations were carried out to investigate scenarios where the structure is primarily to the side or below the robot. The main motivation behind using a fixed monocular forward camera is to employ the proposed methodology for \rebuttal{Commercial Off-The-Shelf (COTS)} vehicles such as the Blue Robotics BlueROV2~\cite{bluerov} that have limited acoustic or other sensing capabilities beyond a camera. 

\begin{figure}[tp]
 \centering
 \begin{subfigure}[b]{0.48\textwidth}
 \centering
 \includegraphics[width=\textwidth]{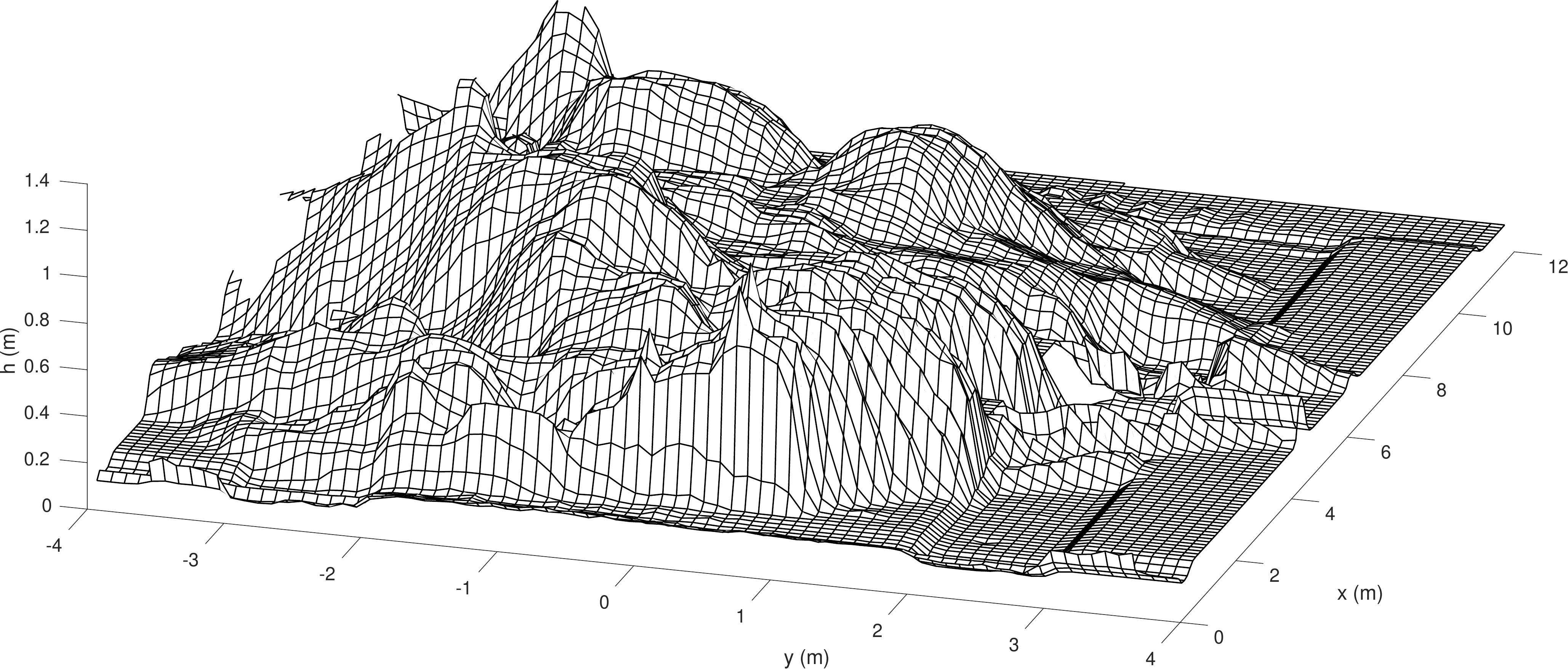}
 \caption{Example of 3D occupancy grid of the simulator terrain used as a ground truth for algorithm evaluation.}
 \label{fig:surface_true_map}
 \end{subfigure}\\
 \begin{subfigure}[b]{0.48\textwidth}
 \centering
 \includegraphics[width=\textwidth]{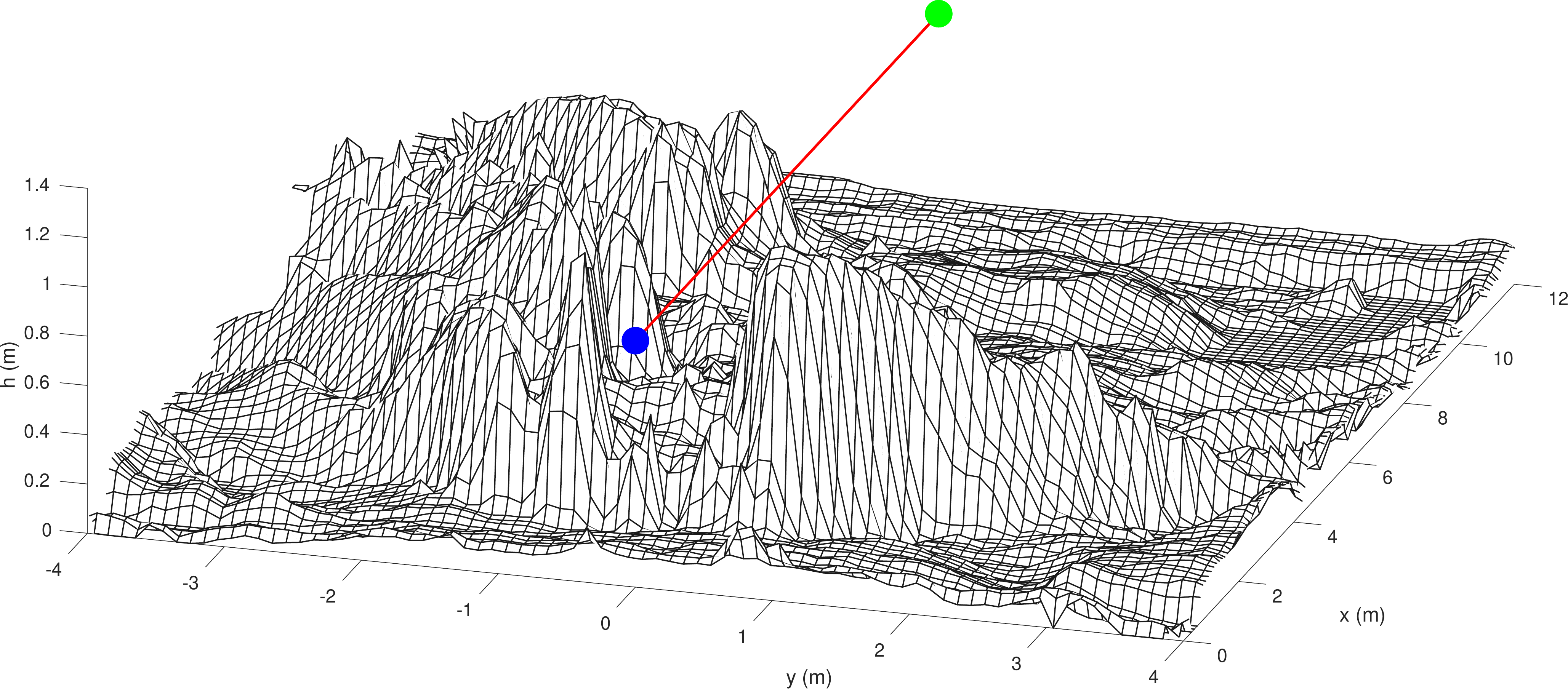}
 \caption{Example of the surface terrain map generated using the proposed approach when accounting for the classification uncertainty. The robot traversed across the transect (red) at a constant $h\!=\!1\mathrm{m}$ from start point (blue marker) to end point (green marker) along the x-axis, while y-axis was set to zero.}
 \label{fig:surface_fused_map}
 \end{subfigure}\\
 \begin{subfigure}[b]{0.48\textwidth}
 \centering
 \includegraphics[width=\textwidth]{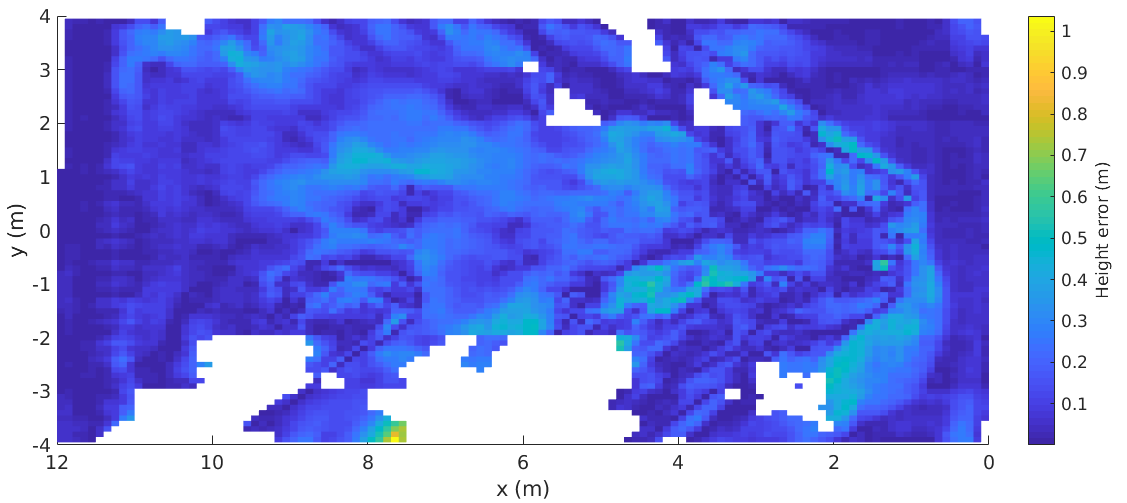}
 \caption{Error between the predicted terrain map compared to the ground truth map (top) to quantify the classifier performance. Note that the white spaces are unobserved regions when using a fixed monocular forward camera.}
 \label{fig:surface_error_map}
 \end{subfigure}
 \caption{Example of the simulation~$2$ scenario for an AUV to follow the transect line in close proximity to the unknown terrain.}
\end{figure}

In both simulations, the robot was commanded to follow a straight line transect from $x\!=\!0\mathrm{m}$ to $x\!=\!12\mathrm{m}$ at constant $h\!=\!1\mathrm{m}$ and $y\!=\!0\mathrm{m}$. In simulation~$1$, the terrain is higher than the AUV traversal height, whereas in the different scene of simulation~$2$ it is below the AUV. The two simulator terrains used as ground truth for algorithm evaluations are shown in Figure~\ref{fig:contour_true_map} and Figure~\ref{fig:surface_true_map} respectively; these were generated from the AUV perspective using a perfect stereo camera model.

Figure~\ref{fig:contour_fused_map} illustrates the terrain prediction for simulation~$1$ using range classification when accounting for the range uncertainty using a monocular camera.  Figure~\ref{fig:contour_error_map} shows the estimated error map by comparing the terrain in Figure~\ref{fig:contour_true_map} and the predicted terrain showing a maximum height error of $\sim\!2.2\mathrm{m}$. This error corresponds to the lack of observation of the ‘gap’ between two terrain structures resulting from the binning of the predicted classes at fixed ranges. This error could potentially be minimized by adding more classes to estimate the ranges at finer distances and/or to employ range estimation methodologies using coarse-to-fine networks~\cite{yhr19}.

\rebuttal{Figure~\ref{fig:Sim_cross_section_comparison} shows the uncertainty associated with the predicted terrain map shown in Figure~\ref{fig:contour_fused_map} in the y-axis at $0\mathrm{m}$ (along the transect), $0.5\mathrm{m}$, and $1\mathrm{m}$. The maximum error between the predicted terrain and the true terrain along the transect is~$\sim\!\!25\mathrm{cm}$. The true terrain remains within the confidence bounds, while robot traversing through the environment, provides us the indication of the use case for close-proximity (i.e. within $0.25$--$1\mathrm{m}$) surveying tasks. In future work we aim to use the confidence bounds as a ‘safety margin’ for local path planning.}

For the simulation~$2$, the AUV was tasked to travel above a different undulating terrain at constant $h\!=\!1\mathrm{m}$. Figure~\ref{fig:surface_fused_map} shows the predicted terrain using a fixed forward monocular camera. The error between the predicted maps, shown in Figure~\ref{fig:surface_error_map}, illustrates the error is minimised in areas along the transect (less than $10\mathrm{cm}$) but can increase on the receding flank of steeper terrain which is less observable to the forward facing camera. 

\begin{figure}
    \centering
    \includegraphics[width=0.48\textwidth]{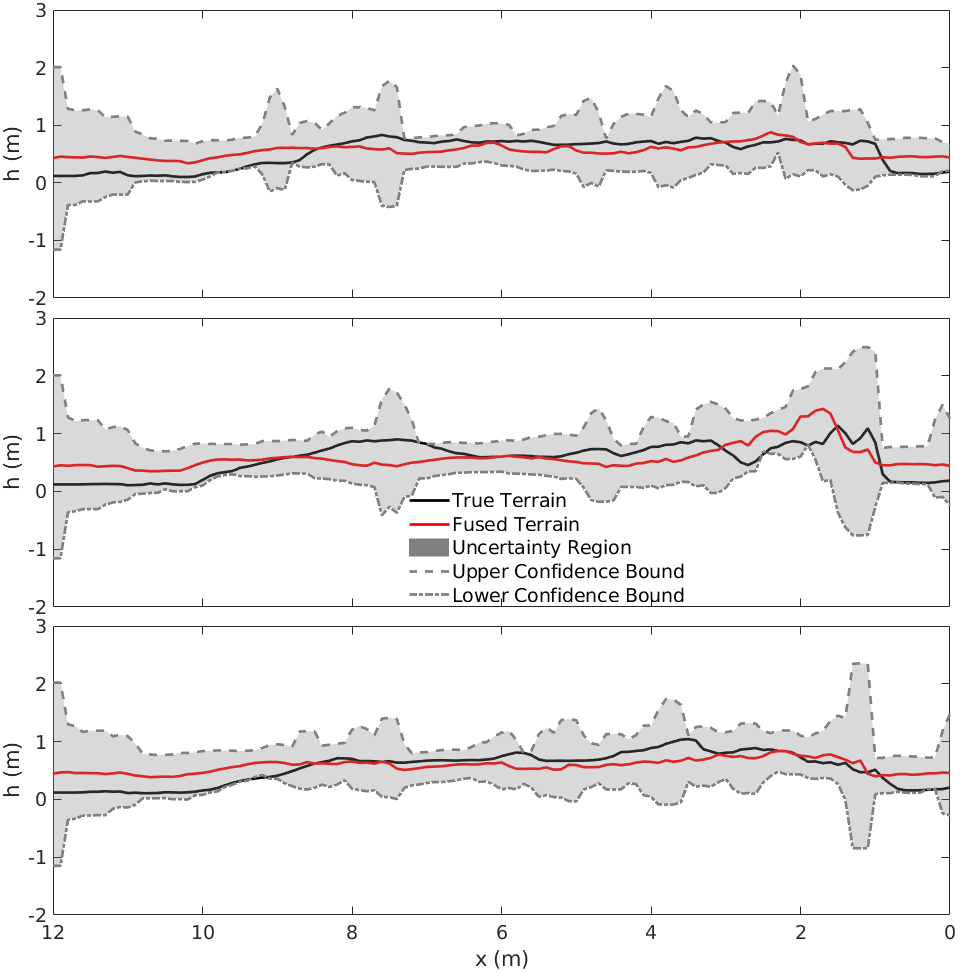}
    \caption{The cross section view of the surface terrain map in the vehicle direction at $y\!=\!0\mathrm{m}$ (Top), at $y\!=\!0.5\mathrm{m}$ (Middle), and at $y\!=\!-0.5\mathrm{m}$ (Lower) in the y-axis showing the comparison between resulting predicted terrain and the true terrain along with the estimated upper and lower confidence bounds. Note that the ground truth offset height has been shifted to the initial robot-centric pose.}
    \label{fig:flyover_cross_section_comparison}
\end{figure}

Figure~\ref{fig:flyover_cross_section_comparison} is the cross sectional view of the terrain shown in Figure~\ref{fig:surface_fused_map} in the robot travelling direction ($y\!=\!0\mathrm{m}$), and at $y\!=\!\pm0.5\mathrm{m}$ of the transect. Figure~\ref{fig:flyover_cross_section_comparison} shows that the maximum error between the predicted terrain, when accounting for the classification uncertainty, and the true terrain in the vehicle traverse direction is $\sim\!30\mathrm{cm}$. 

\begin{figure}
    \centering
    \includegraphics[width=0.48\textwidth]{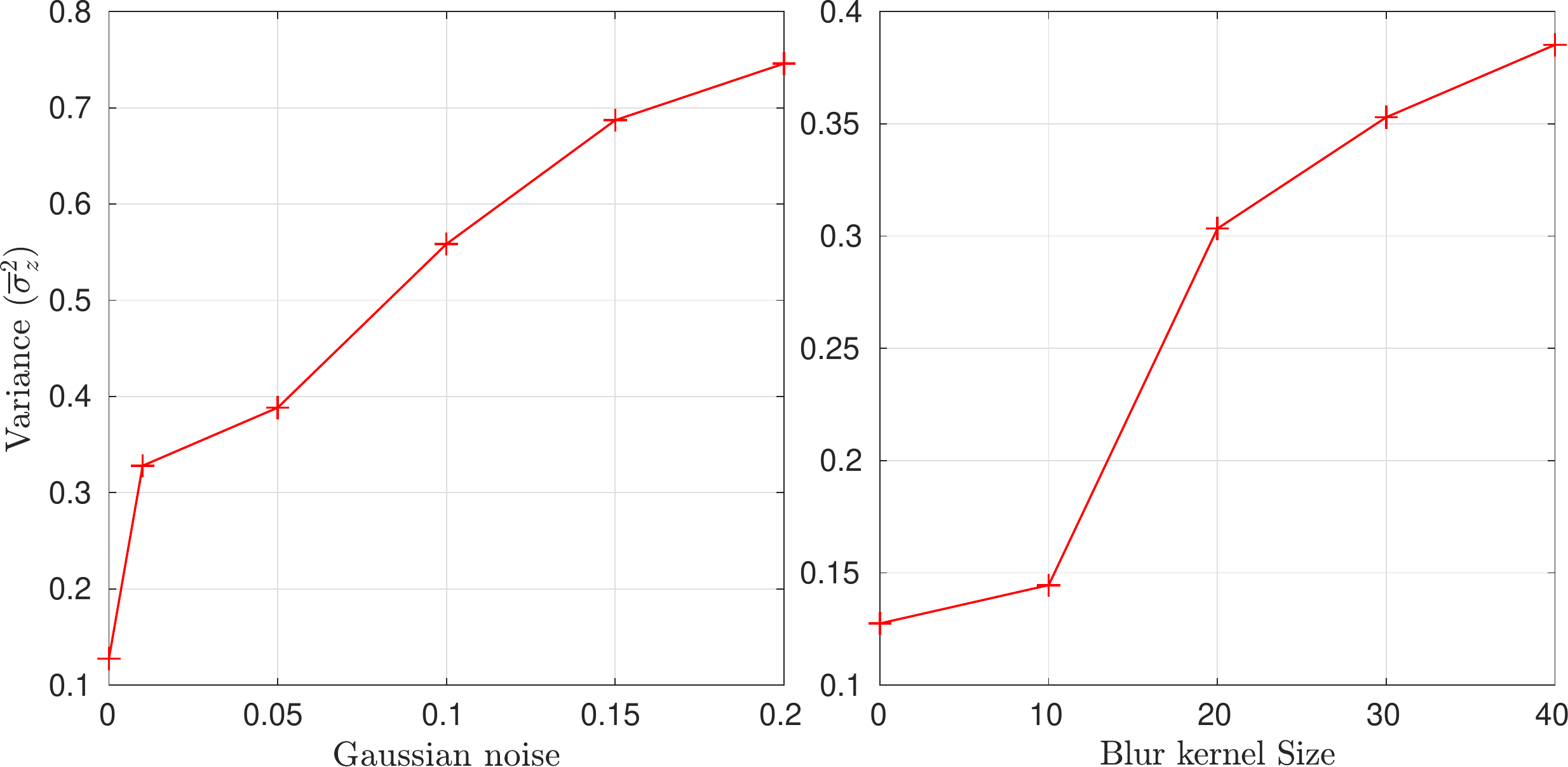}
    \caption{Correlation between the predicted uncertainty and the input image quality. The y-axis ($\overline{\sigma}^{2}_{z}$) represents the mean value of $\sigma^{2}_{z}$ in Eq.~(\ref{eq:range_mean_var}). The x-axis represents the Gaussian white noise with zero mean (left) and blur using kernel size (right).}
    \label{fig:roc_curve}
\end{figure}

\rebuttal{Figure~\ref{fig:roc_curve} shows a correlation between the quality of underwater images and the predicted uncertainty using the classifier. We have gradually increased the Gaussian noise in the input images. Also, we have increased the blur in the input images using increasing value of the kernel size. We then, computed the mean value of $\sigma^{2}_{z}$ in~(\ref{eq:range_mean_var}) per frame. It can be seen that the aleatoric uncertainty estimated by the model monotonically increases with the degradation of the input underwater images.}

\subsection{Experimental Evaluation using Real-World Data}
\label{sec:experiments}
\begin{figure}[!h]
    \centering
    \begin{subfigure}[b]{0.48\textwidth}
         \centering
         \includegraphics[width=\textwidth]{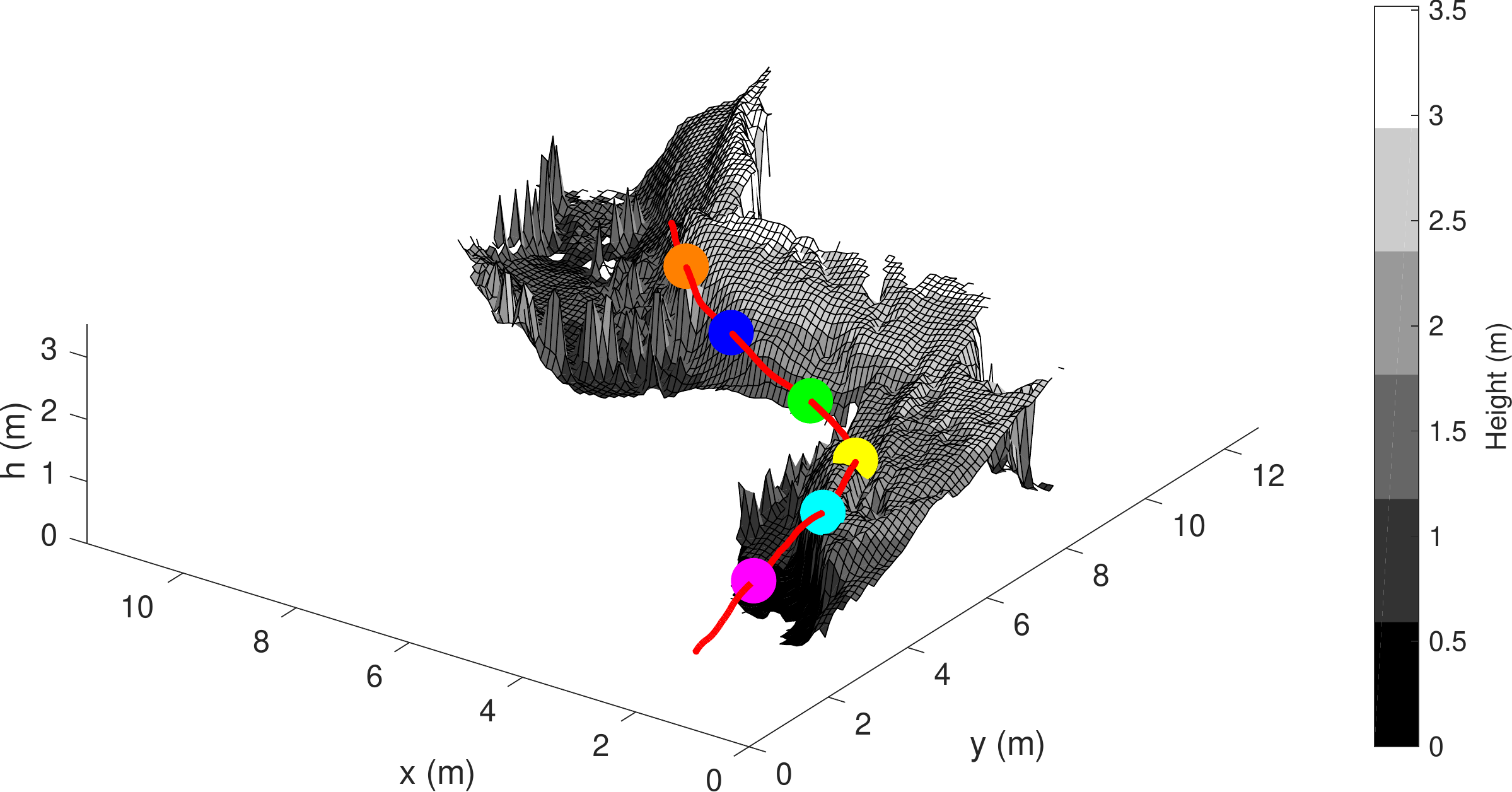}
         \caption{An example of generated terrain map from coral reef survey at John Brewer reef environment. The RangerBot AUV trajectory is shown in red, along with the markers (input images) at different locations along the trajectory.}
         \label{fig:jbr_map}
     \end{subfigure}
     \begin{subfigure}[t]{0.15\textwidth}
         \centering
         \includegraphics[width=\textwidth]{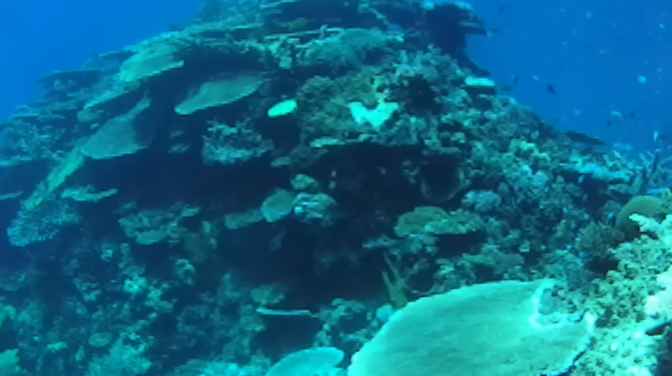}
         \caption{Magenta Marker}
         \label{fig:magenta}
     \end{subfigure}
     \begin{subfigure}[t]{0.15\textwidth}
        \centering
        \includegraphics[width=\textwidth]{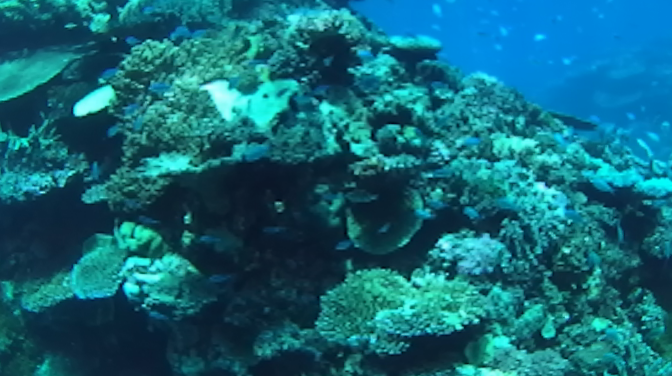}
        \caption{Cyan Marker}
        \label{fig:cyan}
     \end{subfigure}
     \begin{subfigure}[t]{0.15\textwidth}
        \centering
        \includegraphics[width=\textwidth]{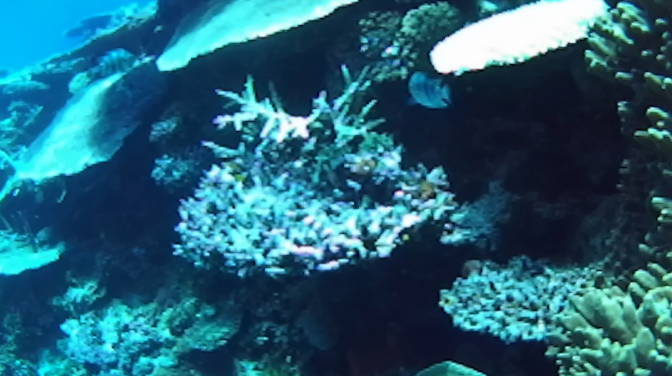}
        \caption{Yellow Marker}
        \label{fig:yellow}
     \end{subfigure}
     
     \begin{subfigure}[t]{0.15\textwidth}
         \centering
         \includegraphics[width=\textwidth]{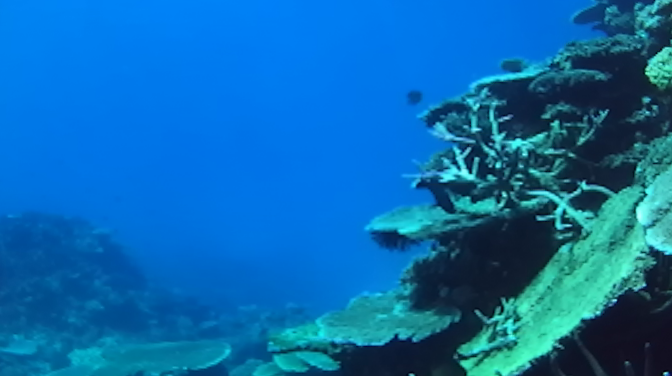}
         \caption{Green Marker}
         \label{fig:green}
     \end{subfigure}
     \begin{subfigure}[t]{0.15\textwidth}
        \centering
        \includegraphics[width=\textwidth]{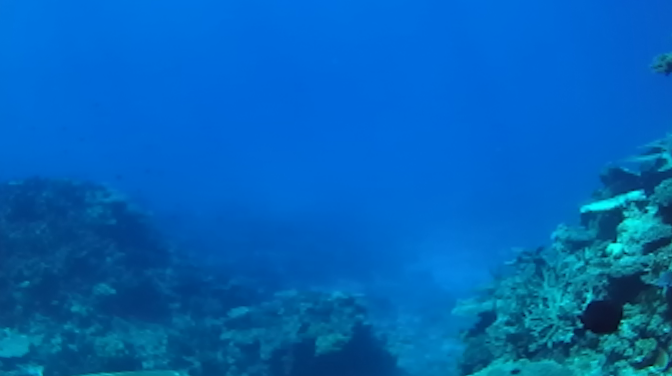}
        \caption{Blue Marker}
        \label{fig:blue}
     \end{subfigure}
     \begin{subfigure}[t]{0.15\textwidth}
        \centering
        \includegraphics[width=\textwidth]{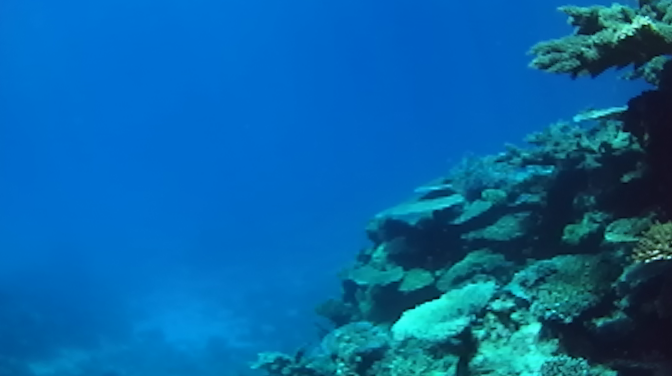}
        \caption{Orange Marker}
        \label{fig:orange}
     \end{subfigure}
     \caption{Results showing the predicted terrain map generated while following the coral reef terrain at John Brewer reef.} 
     \label{fig:jbr_mapping_scenario}
\end{figure}

The performance of the proposed multi-class classifier-based underwater terrain mapping approach was evaluated using image data sets collected by the RangerBot AUV during surveys at John Brewer Reef. \rebuttal{In this analysis, only one of the camera image streams from the RangerBot AUV's forward stereo pair was used to predict the coral reef terrain map.} The predicted terrain map from the field data is shown in Figure~\ref{fig:jbr_mapping_scenario} along with the predicted robot trajectory around the complex reef terrain.

The ability of the proposed multi-class classifier for range prediction with uncertainty can be clearly seen to generate a 3D representation of the scene using a monocular camera for coral reef surveys. To qualitatively evaluate the  performance of the range estimation, example of snapshots (markers) at a regular interval of approximately $8\mathrm{s}$ are shown in Figure~\ref{fig:magenta}--\ref{fig:orange}. As can be seen, the approach accurately reproduces the relative 3D terrain map in front of the camera and reliably predicts the range. These results show promise for producing robust obstacle maps for use in real-time AUV path planning applications, particularly for low-altitude coral surveys. \rebuttal{The mean run-time performance of the proposed network architecture was measured to be $\sim\!8\mathrm{Hz}$ on a Nvidia TX2. The fused map update step performance was almost $50\mathrm{Hz}$ for a map of size $5\!\times5\mathrm{m}$ at $2\mathrm{cm}$ resolution. Faster runtime performance could be achieved by downsampling the point cloud and using the raw map only (without computing the mean height estimated for all the cells).}

\section{Conclusions}
\label{sec:conclusions}
This paper has presented a novel underwater terrain mapping approach using monocular semantic image segmentation. Two DCNNs were trained using simulated and real-world stereo imagery. These models are then used to produce a probabilistic terrain map from images collected during low-altitude coral reef surveys. The prediction output of the multi-class classifier, representing the range to obstacles along with its uncertainty, are transformed to a probability density function of range measurements. The probabilistic range estimate, along with the robot motion update, is then used to generate an estimated height map with confidence bounds. The proposed approach was evaluated using a \rebuttal{underwater} simulated environment, and with field data collected by an AUV. \rebuttal{The simulations and field results show that the proposed approach is feasible for obstacle detection and range estimation using a monocular camera in reef environments, as obstacles are successfully detected within quantified confidence bounds at close proximity ($0.25$--$1\mathrm{m}$) to the robot. Furthermore, it was demonstrated that this algorithm is efficient enough to be run online on standard embedded computing hardware. Future work includes quantitative evaluation of the predicted terrain map along with the confidence bounds for local path planning and collision avoidance algorithms in order to follow the underwater coral reef terrain as closely as possible for surveying tasks.}

\bibliographystyle{IEEEtran}
\bibliography{main}
\end{document}